\pdfoutput=1
\PassOptionsToPackage{x11names}{xcolor}
\documentclass[11pt]{article}

\usepackage[T1]{fontenc}
\usepackage[sc]{mathpazo}
\usepackage{fullpage}

\usepackage[sort&compress,square,numbers]{natbib}
\usepackage{amsthm,amsmath,amssymb}
\usepackage{bm}
\usepackage{comment}
\usepackage[colorinlistoftodos,prependcaption]{todonotes}
\usepackage{xspace}
\usepackage{tikz}
\usetikzlibrary{positioning}
\usepackage{here}
\usepackage{soul}
\usepackage[colorlinks=true,citecolor=Blue4,linkcolor=Firebrick3]{hyperref}

\usepackage[nameinlink]{cleveref}
\usepackage{framed}
\usepackage{authblk}

\newtheorem{theorem}{Theorem}

\newtheorem{corollary}[theorem]{Corollary}
\newtheorem{claim}[theorem]{Claim}
\newtheorem*{claim*}{Claim}

\theoremstyle{definition}
\newtheorem{definition}[theorem]{Definition}
\newtheorem{remark}[theorem]{Remark}

\crefname{theorem}{Theorem}{Theorems}
\crefname{lemma}{Lemma}{Lemmas}
\crefname{claim}{Claim}{Claims}
\crefname{remark}{Remark}{Remarks}
\crefname{observation}{Observation}{Observations}
\crefname{corollary}{Corollary}{Corollaries}
\crefname{appendix}{Appendix}{Appendices}
\crefname{section}{Section}{Sections}
\crefname{equation}{Eq.}{Eqs.}
\crefname{figure}{Figure}{Figures}
\crefname{table}{Table}{Tables}

\newcommand{\rme}{\mathrm{e}}

\newcommand{\bbQ}{\mathbb{Q}}
\newcommand{\bbR}{\mathbb{R}}
\newcommand{\bbZ}{\mathbb{Z}}

\newcommand{\calC}{\mathcal{C}}
\newcommand{\calF}{\mathcal{F}}
\newcommand{\calI}{\mathcal{I}}

\newcommand{\calT}{\mathcal{T}}

\renewcommand{\vec}[1]{\mathbf{\bm{#1}}}
\newcommand{\mat}[1]{\mathbf{\bm{#1}}}

\newcommand{\El}{E_\ell}
\newcommand{\Er}{E_r}
\newcommand{\ZZ}{\mathsf{Z}}
\newcommand{\ZT}{\ZZ_{\mathrm{T}}}
\newcommand{\ZF}{\ZZ_{\mathrm{F}}}

\newcommand{\alg}{\mathcal{A}}

\DeclareMathOperator{\nnz}{\mathop{\mathrm{nz}}}

\newcommand{\cP}{\textup{\textsf{P}}\xspace}
\newcommand{\NP}{\textup{\textsf{NP}}\xspace}
\newcommand{\shP}{\textup{$\sharp$\textsf{P}}\xspace}

\title{Spanning Tree Constrained Determinantal Point Processes are Hard to (Approximately) Evaluate}
\author[1]{Tatsuya Matsuoka\thanks{\href{mailto:ta.matsuoka@nec.com}{\texttt{ta.matsuoka@nec.com}}}}
\author[1]{Naoto Ohsaka\thanks{Corresponding author, \href{mailto:ohsaka@nec.com}{\texttt{ohsaka@nec.com}}}}
\affil[1]{NEC Corporation}
\date{\vspace{-5ex}}

\begin{document}

\maketitle

\begin{abstract}
We consider determinantal point processes (DPPs) constrained by spanning trees.
Given a graph $G=(V,E)$ and a positive semi-definite matrix $\mat{A}$ indexed by $E$,
a \emph{spanning-tree DPP} defines a distribution
such that we draw $S\subseteq E$ with probability proportional to $\det(\mat{A}_S)$ \emph{only if} $S$ induces a spanning tree.
We prove \shP-hardness of computing the normalizing constant for spanning-tree DPPs and
provide an approximation-preserving reduction
from the mixed discriminant, for which FPRAS is not known.
We show similar results for DPPs constrained by \emph{forests}.
\end{abstract}

\section{Introduction}
For a positive semi-definite matrix $\mat{A} \in \bbR^{m \times m}$,
a \emph{determinantal point process (DPP)}~\cite{macchi1975coincidence,borodin2005eynard}
is defined as a probability distribution on the power set $2^{[m]}$,
whose probability mass for each subset $S \subseteq [m]$ is
proportional to $\det(\mat{A}_{S})$, the principal minor of $\mat{A}$.
Originally developed as a model of fermions by \citet{macchi1975coincidence},
DPPs have attracted a great deal of attention from researchers in the machine learning community
because they capture \emph{negative correlations}
and offer a diverse high-quality subset of items.
Applications of DPPs include
image search \cite{kulesza2011kdpps},
video summarization \cite{gong2014diverse}, and
object retrieval \cite{affandi2014learning} to name a few.

One appealing property of DPPs is that
evaluating the \emph{normalizing constant} (a.k.a.~\emph{partition function}),
i.e., $\sum_{S \subseteq [m]} \det(\mat{A}_S)$,
is computationally tractable.
Specifically, the normalizing constant has a closed-form expression $ \det(\mat{A} + \mat{I}) $ \cite{kulesza2012determinantal},
which can be computed by Gaussian elimination in polynomial time~\cite{edmonds1967systems}.
Such tractability is crucial in performing (exact) probabilistic inference efficiently; e.g.,
the probability mass for each subset $S \subseteq [m]$ is obtained by
$ \det(\mat{A}_S) / \det(\mat{A} + \mat{I}) $.
See, e.g., the survey of \citet{kulesza2012determinantal} for more details on probabilistic inference on DPPs.

Since the introduction of DPPs in the machine learning community,
significant effort has been made to express complex distributions
with the imposition of \emph{constraints} on DPPs,
as suggested in the survey \cite[7.3 Research Directions]{kulesza2012determinantal}.
More specifically,
for a set family $\calC \subseteq 2^{[m]}$ representing a certain constraint,
the \emph{$\calC$-constrained DPP} defines a distribution, in which
the probability mass for each subset $S \subseteq 2^{[m]}$ is
proportional to $ \det(\mat{A}_S) \cdot [\![S \in \calC]\!] $,
which is nonzero \emph{only if} $S \in \calC$,
where $[\![S \in \calC]\!]$ is $1$ if $S \in \calC$ and $0$ otherwise.
The corresponding normalizing constant
is thus equal to $\sum_{S \in \calC} \det(\mat{A}_S)$.
The case of $\calC = 2^{[m]}$ coincides with (unconstrained) DPPs.
\citet{kulesza2011kdpps} study the case when
$ \calC $ consists of the size-$k$ subsets,
i.e., the bases of a \emph{uniform matroid}, which is called $k$-DPPs.
Given the eigenvalues of $\mat{A}$,
we can compute the normalizing constant for $k$-DPPs and thus perform probabilistic inference efficiently.
\citet*{celis2017complexity} investigate the case when $ \calC $ consists of the bases of a \emph{partition matroid}, which is called $P$-DPPs~\cite{celis2018fair}.
The normalizing constant for $P$-DPPs
is \shP-hard to compute in general
but is computable in polynomial time if an input partition
of the ground set $[m]$
consists of a constant number of parts \cite{celis2017complexity}.
\citet{celis2017complexity} also examine
\emph{budget constraints}, where
there is a cost vector $\vec{c} \in \bbZ^m$ and
$\calC$ contains any subset $S$ whose cost defined as
$\sum_{i \in S} c_i$ is at most a budget $B \in \bbZ$; i.e.,
$\calC = \{ S \subseteq [m] \mid \sum_{i \in S}c_i \leq B \}$.
The normalizing constant for budget-constrained DPPs can be computed in time
polynomial in $m$ and $\|\vec{c}\|_1$.

In this letter, we consider \emph{spanning-tree constraints} and \emph{forest constraints}.
Recall that for an undirected graph,
a \emph{spanning tree} is a subgraph that connects all vertices and contains no cycles, and
a \emph{forest} is a subgraph that contains no cycles.
Let $ G = (V,E) $ be a simple, undirected graph, and
$ \mat{A} \in \bbR^{E \times E} $ be a positive semi-definite matrix
indexed by the edges of $E$.
We denote the family of the edge sets of all spanning trees of $G$ by $ \calT $ and
the family of the edge sets of all forests of $G$ by $ \calF $; in other words,
$\calT$ and $\calF$ are the families of bases and independent sets of
a \emph{graphic matroid} derived from $G$, respectively.
$\calT$-constrained DPPs and $\calF$-constrained DPPs are then referred to as \emph{spanning-tree DPPs} and \emph{forest DPPs}, respectively.
Sampling spanning trees has several applications, such as 
network centrality \cite{hayashi2016efficient} and graph sparsification \cite{fung2011general}, and
spanning-tree DPPs enable to express negative correlations among the edges of a graph.
Since graphic matroids coincide with neither uniform matroids nor partition matroids,
spanning-tree DPPs could express a different class of probability distributions from both $k$-DPPs and $P$-DPPs.
Hereafter, we
denote by $\ZT$ and $\ZF$
the normalizing constant for spanning-tree DPPs and forest DPPs, respectively;
namely,
\begin{align*}
    \ZT(\mat{A}, G) = \sum_{S \in \calT} \det(\mat{A}_S) \; \text{and} \;
    \ZF(\mat{A}, G) = \sum_{S \in \calF} \det(\mat{A}_S).
\end{align*}

Our objective in this paper is to
investigate the computational complexity of estimating $\ZT$ and $\ZF$.
In the special case that $\mat{A}$ is an identity matrix $\mat{I}$,
$\ZT(\mat{I},G)$ corresponds to the number of spanning trees in $G$ and
$\ZF(\mat{I},G)$ corresponds to the number of forests in $G$.
We can count the number of spanning trees in a graph
in polynomial time using Kirchhoff's matrix-tree theorem \cite{kirchhoff1847ueber}.
On the other hand, it is already \shP-hard to count the number of forests,
even if an input graph $G$ is restricted to be bipartite and planar \cite{vertigan1992computational}, while
there is an FPRAS for $\ZF(\mat{I},G)$ when $G$ is a dense graph \cite{annan1994randomised}.
Note also that the Tutte polynomial,
which includes the number of forests $\ZF(\mat{I},G)$ as a special case at point $(2,1)$,
can be computed in polynomial time if $G$ has constant treewidth \cite{noble1998evaluating,andrzejak1998algorithm}.

\subsection{Our Contributions}

\paragraph{\shP-hardness of Computing $\ZT$ and $\ZF$.}
We prove that it is \shP-hard to compute
the normalizing constants for
spanning-tree DPPs $\ZT(\mat{A},G)$ and
forest DPPs $\ZF(\mat{A},G)$ for a graph $G=(V,E)$ and a positive semi-definite matrix
$\mat{A} \in \bbQ^{E \times E}$.
The \shP-hardness result still holds even when both of $G$ and $\mat{A}$
are restricted to have treewidth $2$,
which is in contrast to the fact that
$\ZT(\mat{I},G)$ and $\ZF(\mat{I},G)$ can be computed in polynomial time for bounded-tree graph $G$.
Here, the \emph{treewidth} of an $m \times m$ matrix $\mat{A}$
is defined as the treewidth of the graph $ ([m], \nnz(\mat{A})) $, where
$ \nnz(\mat{A}) = \{ (i, j) \mid A_{i,j} \neq 0, i \neq j \} $ (see, e.g., \cite{courcelle2001fixed}).
The proofs of \cref{thm:st-hard,cor:fo-hard} are
provided in \cref{sec:st-hard,sec:fo-hard}, respectively.
In particular, we present a polynomial-time reduction
from the number of all perfect matchings in a ($3$-regular) bipartite graph to $\ZT$ in the proof of \cref{thm:st-hard}.

\begin{theorem}
\label{thm:st-hard}
Let $G=(V,E)$ be a simple, undirected graph,
$\mat{A} \in \bbQ^{E \times E}$ be a positive semi-definite matrix, and
$\calT$ be the family of the edge sets of all spanning trees of $G$.
Then, it is \shP-hard to compute $ \ZT(\mat{A},G) = \sum_{S \in \calT} \det(\mat{A}_S) $ exactly.
The same hardness holds even if
$\mat{A}$ is a $(0,1)$-matrix of treewidth $2$, and
$G$ is of treewidth $2$.
\end{theorem}

\begin{corollary}
\label{cor:fo-hard}
Let $G=(V,E)$ be a simple, undirected graph,
$\mat{A} \in \bbQ^{E \times E}$ be a positive semi-definite matrix, and
$\calF$ be the family of the edge sets of all forests of $G$.
Then, it is \shP-hard to compute $ \ZF(\mat{A}, G) = \sum_{S \in \calF} \det(\mat{A}_S) $ exactly even if
$\mat{A}$ is of treewidth $2$, and $G$ is of treewidth $2$.
\end{corollary}

\begin{remark}
Treewidth \cite{robertson1986graph} is one of the most fundamental graph-theoretic parameters,
measuring the ``treelikeness'' of a graph;
e.g., trees have treewidth $1$,
series-parallel graphs have treewidth at most $2$,
$n$-vertex planar graphs have treewidth $\mathcal{O}(\sqrt{n})$, and
$n$-cliques have treewidth $n-1$.
Many \NP-hard problems on graphs have been shown to be
polynomial-time solvable for bounded-treewidth graphs,
see, e.g., \cite{cygan2015parameterized}.
In particular,
the number of forests in a graph is computable in polynomial time if the treewidth is a constant \cite{noble1998evaluating,andrzejak1998algorithm}.
Our results, however, refute the possibility of
such an efficient algorithm for bounded-tree graphs
(unless \cP $=$ \shP).
\end{remark}

\paragraph{Approximation-Preserving Reduction from Mixed Discriminant.}
Beyond the difficulty regarding exact computation,
we analyze \emph{approximability} of $\ZT$ and $\ZF$.
We stress that \shP-hardness for a particular problem does not necessarily rule out
the existence of efficient approximation algorithms for it; e.g.,
the number of perfect matchings in a bipartite graph can be approximated within an arbitrary precision \cite{jerrum2004polynomial},
though it is a \shP-complete problem \cite{valiant1979complexity}.

Here, we introduce several definitions regarding approximate computing.
We say that an estimate $\hat{\ZZ}$ is
a \emph{$\rho$-approximation} to the true value $\ZZ$ for $\rho \geq 1$ if it holds that
\begin{align*}
(1/\rho) \cdot \ZZ \leq \hat{\ZZ} \leq \rho \cdot \ZZ.
\end{align*}
We then define a fully polynomial-time randomized approximation scheme.
\begin{definition}
For a function $f: \Sigma^* \to \bbR$,
a \emph{fully polynomial-time randomized approximation scheme (FPRAS)}
is a randomized algorithm $\alg$ that takes
an instance $x \in \Sigma^*$ of $f$ and an error tolerance $\epsilon \in (0,1)$ as input and
satisfies the following conditions:
\begin{itemize}
    \item For every $x \in \Sigma^*$ and $\epsilon \in (0,1)$,
    $\alg$ outputs an $\rme^{-\epsilon}$-approximation to $f(x)$ with probability at least $\frac{3}{4}$; i.e.,
    \begin{align}
    \label{eq:fpras}
        \Pr_{\alg}\Bigl[ \rme^{-\epsilon} \cdot f(x) \leq \alg(x) \leq \rme^{\epsilon} \cdot f(x) \Bigr] \geq \frac{3}{4},
    \end{align}
    where $\alg(x)$ denotes $\alg$'s output on $x$.\footnote{Note that the constant $\frac{3}{4}$ in \cref{eq:fpras} can be replaced by
    any number in $(\frac{1}{2},1)$ \cite{jerrum1986random}.}
    \item The running time of $\alg$ is bounded by
    a polynomial in $|x|$ and $\epsilon^{-1}$,
    where $|x|$ denotes the number of bits required for representing $x$.
\end{itemize}
\end{definition}

We finally define the notion of approximation-preserving reduction 
according to \citet*{dyer2004relative},
which can be used to translate an FPRAS for a function $g$ into an FPRAS for another function $f$.

\begin{definition}
For two functions
$f: \Sigma^* \to \bbR$ and
$g: \Sigma^* \to \bbR$,
an \emph{approximation-preserving reduction (AP-reduction)}
from $f$ to $g$
is a randomized algorithm $\alg$ that takes
an instance $x \in \Sigma^*$ of $f$ and an error tolerance $\epsilon \in (0,1)$ as input and
satisfies the following conditions:
\begin{itemize}
    \item Every oracle call for $g$
    made by $\alg$ is of the form $(y,\delta)$,
    where $y \in \Sigma^*$ is an instance of $g$ and
    $\delta \in (0,1)$ is an error tolerance satisfying
    that $ \delta^{-1} $ is bounded by a polynomial in
    $|x|$ and $\epsilon^{-1}$.
    \item If the oracle meets the specification for an FPRAS for $g$, then $\alg$ meets the specification for an FPRAS for $f$.
    \item The running time of $\alg$ is bounded by
    a polynomial in $|x|$ and $\epsilon^{-1}$.
\end{itemize}
We say that $f$ is \emph{AP-reducible} to $g$ if an AP-reduction from $f$ to $g$ exists.
\end{definition}
It is known \cite{dyer2004relative} that
assuming $f$ to be AP-reducible to $g$,
an FPRAS for $g$ implies an FPRAS for $f$; in other words,
if $f$ does not admit an FPRAS (under some plausible assumption), 
then $g$ does not as well.

Our technical results are AP-reductions from the mixed discriminant to $\ZT$ and $\ZF$.
Here, the \emph{mixed discriminant} for $n$ positive semi-definite matrices
$\mat{K}^1, \ldots, \mat{K}^n \in \bbR^{n \times n}$ is defined as follows:
\begin{align*}
    D(\mat{K}^1, \ldots, \mat{K}^n) = \frac{\partial^n}{\partial x_1 \cdots \partial x_n} \det(x_1 \mat{K}^1 + \cdots + x_m \mat{K}^n).
\end{align*}

\begin{theorem}
\label{thm:st-fpras}
The mixed discriminant $D$ is AP-reducible to $\ZT$; therefore,
if there exists an FPRAS for $\ZT$,
then there exists an FPRAS for $D$.
\end{theorem}

\begin{theorem}
\label{thm:fo-fpras}
The mixed discriminant $D$ is AP-reducible to $\ZF$; therefore,
if there exists an FPRAS for $\ZF$,
then there exists an FPRAS for $D$.
\end{theorem}

Because an FPRAS for the mixed discriminant has not been known and
its existence is suspected to be false \cite{gurvits2005complexity},
our AP-reductions give evidence that
$\ZT$ and $\ZF$ are unlikely to admit an FPRAS.
Furthermore,
due to the equivalence between (approximate) counting and (approximate) sampling
(cf.~\cite[Section B]{celis2017complexity}),
we can immediately rule out the existence of polynomial-time sampling algorithms
for spanning-tree DPPs and forest DPPs
(unless the mixed discriminant admits an FPRAS).
The proofs of \cref{thm:st-fpras,thm:fo-fpras} are
provided in \cref{sec:st-fpras,sec:fo-fpras}, respectively.
It should be noted that we can no longer use polynomial interpolation as used in the proof of \cref{cor:fo-hard}, which does not preserve the closeness of approximation.

\section{Proof of \cref{thm:st-hard}}
\label{sec:st-hard}

\begin{figure*}[tbp]
\centering
\scalebox{0.9}{
\begin{tikzpicture}
[circlenode/.style={draw, circle, minimum height=1cm}]
\node[circlenode](v1){$u_1$};
\node[draw, rectangle, minimum width=1.5cm, right=1cm of v1](v2){$u_1w_{j_{1,2}}$};
\node[draw, rectangle, minimum width=1.5cm, above=1cm of v2](v3){$u_1w_{j_{1,1}}$};
\node[draw, rectangle, minimum width=1.5cm, below=1cm of v2](v4){$u_1w_{j_{1,3}}$};
\node[circlenode, right=1cm of v2](v5){$u_2$};
\node[draw, rectangle, minimum width=1.5cm, right=1cm of v5](v6){$u_2w_{j_{2,2}}$};
\node[draw, rectangle, minimum width=1.5cm, above=1cm of v6](v7){$u_2w_{j_{2,1}}$};
\node[draw, rectangle, minimum width=1.5cm, below=1cm of v6](v8){$u_2w_{j_{2,3}}$};
\node[circlenode, right=1cm of v6](v9){$u_3$};
\draw node(vcdots)[right=0.5cm of v9]{$\cdots$};
\node[circlenode, right=0.5cm of vcdots](v10){$u_n$};
\node[draw, rectangle, minimum width=1.5cm, right=1cm of v10](v11){$u_nw_{j_{n,2}}$};
\node[draw, rectangle, minimum width=1.5cm, above=1cm of v11](v12){$u_nw_{j_{n,1}}$};
\node[draw, rectangle, minimum width=1.5cm, below=1cm of v11](v13){$u_nw_{j_{n,3}}$};
\node[circlenode, right=1cm of v11](v14){$u_{n+1}$};

\foreach \u / \v in {v1/v2, v1/v4, v5/v7, v5/v8, v10/v11, v10/v12}
    \draw (\u)--(\v);
\foreach \u / \v in {v1/v3, v5/v6, v10/v13}
    \draw[ultra thick] (\u)--(\v);
\foreach \u / \v in {v2/v5, v3/v5, v4/v5, v6/v9, v7/v9, v8/v9, v11/v14, v12/v14, v13/v14}
    \draw[ultra thick, dashed] (\u)--(\v);
\end{tikzpicture}
}
\caption{Construction of $G=(V,E)$ from a bipartite graph $B=(U, W; F)$
in the proof of \cref{thm:st-hard}.
Solid lines are edges of $\El$ and dashed lines are edges of $\Er$.
If the set $S \subseteq E$ of bold edges induces a spanning tree of $G$ and
$\det(\mat{A}_S) \neq 0$, then
$S$ must include $\Er$ while $S \cap \El$ forms a perfect matching in $B$.
}
\label{fig:constr}
\end{figure*}
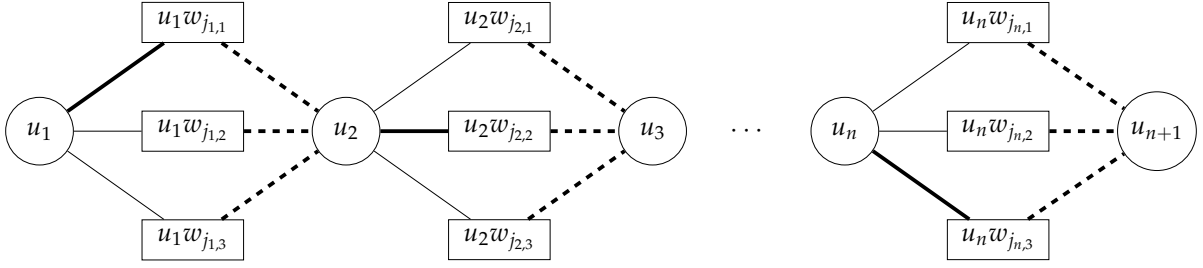

We show a polynomial-time (many-one) reduction from the problem of counting the number of all perfect matchings in a bipartite graph,
which is \shP-complete \cite{valiant1979complexity}.
Let $ B = (U, W; F) $ be a bipartite graph,
where $U = \{u_1, \ldots, u_n\}$, $W=\{w_1, \ldots, w_n\}$, and
$ F \subseteq U \times W $ is a set of $m$ edges between $U$ and $W$.
A \emph{perfect matching} of $B$ is a set of $n$ edges in $F$ that are pairwise vertex-disjoint.
Given $B$,
we first construct a simple, undirected graph $ G=(V,E) $
such that
$ V = \{ u_1, \ldots, u_n, u_{n+1} \} \cup \{ u_iw_j \mid (u_i,w_j) \in F \} $,
where $u_{n+1}$ is a dummy vertex not in $U$, and
$ E = \El \cup \Er $, where
$ \El = \{ (u_i, u_iw_j) \mid (u_i, w_j) \in F \} $ and
$ \Er = \{ (u_iw_j, u_{i+1}) \mid (u_i, w_j) \in F \} $.
See \cref{fig:constr} for an example.
Note that $ |V| = n+m+1 $ and $|E| = 2m$.
We then construct
a $(0,1)$-matrix $\mat{A} \in \{0,1\}^{E \times E}$ as
\begin{align*}
    \mat{A} =
    \begin{bmatrix}
    \mat{A}' & \mat{O}^\top \\
    \mat{O} & \mat{I}
    \end{bmatrix},
\end{align*}
where
$\mat{O}$ is an $ \Er \times \El $ all-zero matrix,
$\mat{I}$ is an $\Er\times \Er$ identity matrix, and
$ \mat{A}'$ is an $\El \times \El$ $(0,1)$-matrix defined as follows:
\begin{align*}
    A'_{(u_{i_1}, u_{i_1}w_{j_1}), (u_{i_2}, u_{i_2}w_{j_2})} = 
    \begin{cases}
    1 & \text{if } w_{j_1} = w_{j_2}, \\
    0 & \text{otherwise}.
    \end{cases}
\end{align*}
By definition, $\mat{A}$ is positive semi-definite.
It is easy to observe that
$ \det(\mat{A}_{S}) $ for $S \subseteq E$ is
1 if $S$ includes no pair of two edges
$ (u_{i_1}, u_{i_1} w_j) $ and $ (u_{i_2}, u_{i_2} w_j) $
for any $j$ and distinct $i_1, i_2$, and
0 otherwise.

We now use the following claim to ensure that
$ \ZT(\mat{A},G) $ is equal to the number of all perfect matchings in $B$, which completes the correctness of the reduction.

\begin{claim}
\label{claim:thm:st-hard}
For an edge set $S \subseteq E$,
$ \det(\mat{A}_{S}) \neq 0 $ and $S$ induces a spanning tree of $G$
if and only if
it holds that $\Er \subseteq S$ and
the edge set $ \{ (u_i, w_j) \mid (u_i, u_i w_j) \in S \cap \El \} $
is a perfect matching of $B$.
\end{claim}

\begin{proof}[Proof of \cref{claim:thm:st-hard}]
Assume that for $S \subseteq E$,
$ \det(\mat{A}_{S}) \neq 0 $ and $S$ induces a spanning tree of $G$.
We first observe that $\Er \subseteq S$;
otherwise, $S$ contains at least $n+1$ edges in $\El$
since $ |S|=|V|-1 = n+m$ and $|\Er| = m$,
resulting in that $\det(\mat{A}_{S}) = 0$.
Then, denote $M = S \cap \El$ with $|M|=n$.
Since $\det(\mat{A}_{S}) \neq 0$,
$S$ includes no pair of two edges $ (u_{i_1}, u_{i_1} w_j) $ and $ (u_{i_2}, u_{i_2} w_j) $
for any $j$ and distinct $ i_1, i_2 $.
Since $S$ induces a spanning tree,
$S$ includes no pair of two edges $ (u_i, u_i w_{j_1}) $ and $ (u_i, u_i w_{j_2}) $
for any $i$ and distinct $ j_1, j_2 $;
otherwise, such a pair and edges in $\Er$ form a cycle:
$ (u_i, u_i w_{j_1})$,
$(u_i w_{j_1}, u_{i+1})$ ($\in \Er \subseteq S$),
$(u_{i+1}, u_i w_{j_2})$,
$(u_i w_{j_2}, u_i)$ ($\in \Er \subseteq S $).
Consequently, the edge set
$ \{ (u_i, w_j) \mid (u_i, u_i w_j) \in M \} $
should be a perfect matching of $G$.
The converse direction is obvious.
\end{proof}

We finally show the restricted-case \shP-hardness.
Observe that the treewidth of $G$ is $2$ 
because $G$ is a series-parallel graph (but not a tree)~\cite{bodlaender1998partial}.
Let $B$ be a $3$-regular bipartite graph, for which
counting the number of all perfect matchings is \shP-complete~\cite[Theorem 6.2]{dagum1992approximating}.
It turns out that the graph formed by the nonzero entries of $\mat{A}$
is the union of $3$-cliques, each of which has treewidth $2$.
Hence $\mat{A}$ has treewidth $2$.
\qed

\section{Proof of \cref{cor:fo-hard}}
\label{sec:fo-hard}

We show a polynomial-time (Turing) reduction
from the normalizing constant for spanning-tree DPPs to that for forest DPPs,
which has been proven to be \shP-hard above.
Let $G = (V,E)$ be a simple, undirected graph of treewidth $2$, where $n=|V|$ and $m=|E|$, and
$\mat{A} \in \bbQ^{E \times E}$ be a positive semi-definite $(0,1)$-matrix of treewidth $2$.
For a positive integer $x$,
let us consider a forest DPP defined by $x\mat{A}$ and $G$.
Let $\calF$ denote the family of the edge sets of all forests of $G$ and
$\calT$ denote the family of the edge sets of all spanning trees of $G$.
We define a polynomial $\ZZ$ in $x$ as follows:
\begin{align*}
    \ZZ(x) = \ZF(x\mat{A},G) = \sum_{S \in \calF} \det((x\mat{A})_S).
\end{align*}
Note that the degree of $\ZZ$ is at most $n-1$.
Since an edge set $S \subseteq E$ induces a spanning tree if and only if $S \in \calF$ and $|S|=n-1$,
we can expand $\ZZ(x)$ as follows:
\begin{align*}
    \ZZ(x) & = \sum_{S \in \calF} \det((x\mat{A})_S) \\
    & = \sum_{\substack{S \in \calF \\ |S| < n-1}} x^{|S|} \det(\mat{A}_S)
    + \sum_{\substack{S \in \calF \\ |S| = n-1}} x^{n-1} \det(\mat{A}_S) \\
    & = \sum_{0 \leq k < n-1} \alpha_k x^k + \left(\sum_{S \in \calT} \det(\mat{A}_S) \right) x^{n-1},
\end{align*}
where $ \alpha_k $ for $0 \leq k < n-1$ is some coefficient.
Given $\ZZ(1), \ZZ(2), \ldots, \ZZ(n)$,
each of which is the normalizing constant for a forest DPP,
we can recover $\sum_{S \in \calT} \det(\mat{A}_S)$ by Lagrange interpolation as desired,
which completes the reduction.
Note that the matrix $x\mat{A}$ is positive semi-definite and has treewidth $2$ for any $x > 0$.
\qed

\section{Proof of \cref{thm:st-fpras}}
\label{sec:st-fpras}
We construct an AP-reduction from the mixed discriminant $D$ to $\ZT$.
Suppose we have an FPRAS for $\ZT$.
Let $\epsilon \in (0,1)$ be an error tolerance for $D$;
i.e., we are asked to estimate $D$ within a factor of $\rme^{\epsilon}$.

Let $\mat{K}^1, \ldots, \mat{K}^n$ be $n$ positive semi-definite matrices in $\bbQ^{n \times n}$, and let $m = n^2$.
According to \cite[Proof of Lemma 12]{celis2017complexity},
we construct an $m \times m$ positive semi-definite matrix $\mat{A}$ and
an equal-sized partition of $[m]$, denoted
$P_1, P_2, \ldots, P_n$ with $ |P_1|=|P_2|= \cdots = |P_n|=n $,
such that the following is satisfied:
\begin{align*}
    \sum_{S \in \calC} \det(\mat{A}_{S}) = m! \; D(\mat{K}^1, \ldots, \mat{K}^n),
\end{align*}
where
$ \calC = \{S \in {[m] \choose n} \mid |S \cap P_i| = 1\mbox{ for all }i \in [n]\} $.

We then construct a simple, undirected graph $G=(V,E)$ such that
\begin{align}
    V & = \{v_1, \ldots, v_n, v_{n+1}\} \cup \{ w_{i,j} \mid i \in [n], j \in P_i \}, \label{eq:def-G-V}\\
    E & = \El \cup \Er, \text{ where} \label{eq:def-G-E}\\
    \El & = \{ (v_i, w_{i,j}) \mid i \in [n], j \in P_i \}, \text{ and} \label{eq:def-G-El}\\
    \Er & = \{ (w_{i,j}, v_{i+1}) \mid i \in [n],j \in P_i \}. \label{eq:def-G-Er}
\end{align}
Note that $|V| = n+m+1$ and $|E|=2m$.
We further construct a matrix $\mat{B} \in \bbQ^{E \times E}$ defined as follows:
\begin{align}
\label{eq:def-B}
    \mat{B} =
    \begin{bmatrix}
    \mat{A}' & \mat{O}^\top \\
    \mat{O} & \mat{I}
    \end{bmatrix},
\end{align}
where
$\mat{O}$ is an $ \Er \times \El $ all-zero matrix,
$\mat{I}$ is an $\Er\times \Er$ identity matrix, and
$ \mat{A}'$ is an $\El \times \El$ matrix defined as
$A'_{(v_{i_1}, w_{i_1, j_1}), (v_{i_2}, w_{i_2,j_2})} = A_{j_1, j_2}$
for $(v_{i_1}, w_{i_1, j_1}), (v_{i_2}, w_{i_2,j_2}) \in \El$.
By definition, $\mat{B}$ is positive semi-definite.
Let $\calT$ be the family of the edge sets of all spanning trees of $G$.
We then claim the following.

\begin{claim}
\label{claim:thm:st-fpras}
Let $S$ be an edge set with $\Er \subseteq S \subseteq E$.
Then, $S$ induces a spanning tree of $G$ if and only if 
the set $\{ j \in [m] \mid (v_i, w_{i,j}) \in S \cap \El \}$ is contained in $\calC$.
\end{claim}
\begin{proof}[Proof of \cref{claim:thm:st-fpras}]
Suppose $S$ with $\Er \subseteq S \subseteq E$
induces a spanning tree.
Since $\Er \subseteq S$, for each $i \in [n]$,
$S$ contains exactly one edge $(v_i, w_{i,j}) \in \El$ for some $j \in P_i$
because otherwise, the subgraph induced by $S$ becomes disconnected or has a cycle.
The converse direction is obvious.
\end{proof}
The following equality is a direct consequence of \cref{claim:thm:st-fpras}.
\begin{align}
\label{eq:sum-st}
\sum_{S \in \calT: \Er \subseteq S} \det(\mat{B}_S) =
\sum_{S \in \calC} \det(\mat{A}_S).
\end{align}

Introduce a positive rational number $x \in \bbQ$ (whose value will be determined later) and
define a matrix $\mat{X} \in \bbQ^{E \times E}$ depending on $x$ as follows:
\begin{align*}
    X_{i,j} =
    \begin{cases}
    1 & \text{if } i, j \in \El, \\
    x & \text{if } i \in \El, j \in \Er, \\
    x & \text{if } j \in \El, i \in \Er, \\
    x^2 & \text{if } i,j \in \Er.
    \end{cases}
\end{align*}
Consider the matrix $ \mat{B} \circ \mat{X} $, where
$\circ$ denotes the Hadamard product operator;
namely, $ (\mat{B} \circ \mat{X})_{i,j} = B_{i,j} \cdot X_{i,j}$ for each $i,j \in E$.
Since $\mat{X}$ is positive semi-definite (its eigenvalues are $0$ and $m(x^2+1)$),
so is $\mat{B} \circ \mat{X}$ by the Schur product theorem \cite{schur1911bemerkungen}.
It is easy to show that for each $S \subseteq E$,
\begin{align*}
    \det((\mat{B} \circ \mat{X})_S) = x^{2|S \cap \Er|} \det(\mat{B}_S).
\end{align*}
We define a univariate polynomial $\ZZ$ in $x$ as follows:
\begin{align*}
\ZZ(x) = \ZT(\mat{B} \circ \mat{X}, G) = \sum_{S \in \calT} x^{2|S \cap \Er|} \det(\mat{B}_S).
\end{align*}
The degree of $\ZZ$ is at most $2m$.
By \cref{eq:sum-st}, the coefficient of $x^{2m}$
is exactly equal to the desired value, i.e.,
$ m! \; D(\mat{K}^1, \ldots, \mat{K}^n) $.
We now identify the value of $x$ for which
$\ZZ(x)$ is sufficiently close to  $m! \; D(\mat{K}^1, \ldots, \mat{K}^n)$.
We first verify whether there exists a subset $S \in \calT$ with $\Er \subseteq S$ such that $\det(\mat{B}_S) > 0$
because otherwise, we can safely declare that
\cref{eq:sum-st} is $0$; i.e., 
$D(\mat{K}^1, \ldots, \mat{K}^n)$ is $0$ as well.
Such a subset can be found (if exists) by matroid intersection because
$ \calI_1 = \{ S \mid \det(\mat{B}_S) > 0 \} $ forms a linear matroid and
$ \calI_2 = \{ S \mid S \in \calF, \Er \subseteq S \}$ forms a partition matroid,
where $\calF$ is the family of the edge sets of all forests of $G$.
Denote the subset found by $\tilde{S} \in \calI_1 \cap \calI_2$.
We then fix the value of $x$ as
\begin{align}
\label{eq:x-st-fpras}
x = \frac{\det(\mat{B}+\mat{I})}{\det(\mat{B}_{\tilde{S}})} \frac{2}{\epsilon}.
\end{align}
Observing that $x > 1$, we bound $\ZZ(x)$ from above as follows:
\begin{align*}
    & \sum_{S \in \calT : \Er \subseteq S} x^{2|S \cap \Er|} \det(\mat{B}_S) +
    \sum_{S \in \calT : \Er \not \subseteq S} x^{2|S \cap \Er|} \det(\mat{B}_S) \\
    & \leq \sum_{S \in \calT: \Er \subseteq S} x^{2m} \det(\mat{B}_S) +
    \sum_{S \in \calT: \Er \not \subseteq S} x^{2m-2} \det(\mat{B}_S) \\
    & \leq x^{2m} \sum_{S \in \calT: \Er \subseteq S} \det(\mat{B}_S)
    \left[ 1 + \frac{\sum\limits_{S \in \calT: \Er \not \subseteq S} \det(\mat{B}_S)}{\sum\limits_{S \in \calT: \Er \subseteq S} \det(\mat{B}_S)}\frac{1}{x^2} \right] \\
    & \leq x^{2m} \sum_{S \in \calT: \Er \subseteq S} \det(\mat{B}_S)
    \left[ 1 + \frac{\sum\limits_{\substack{S \in \calT \\ \Er \not \subseteq S}} \det(\mat{B}_S)}{\sum\limits_{\substack{S \in \calT \\ \Er \subseteq S}} \det(\mat{B}_S)} \frac{\det(\mat{B}_{\tilde{S}})}{\det(\mat{B}+\mat{I})} \frac{\epsilon}{2} \frac{1}{x} \right] \\
    & \leq \Bigl(1+\frac{\epsilon}{2}\Bigr) x^{2m} \sum_{S \in \calT: \Er \subseteq S} \det(\mat{B}_S) \\
    & \leq \rme^{\epsilon/2} x^{2m} m! \; D(\mat{K}^1, \ldots, \mat{K}^n).
\end{align*}
Since
$ \ZZ(x) \geq x^{2m} m! \; D(\mat{K}^1, \ldots, \mat{K}^n)$, we have that
\begin{align}
\label{eq:st-fpras-bound}
    D(\mat{K}^1, \ldots, \mat{K}^n) \leq \frac{\ZZ(x)}{x^{2m} m!} \leq \rme^{\epsilon/2}\cdot
    D(\mat{K}^1, \ldots, \mat{K}^n).
\end{align}

We are finally ready to describe the AP-reduction from
the mixed discriminant $D$ to $\ZT$.
\begin{oframed}
\begin{center}
    \textbf{AP-reduction from $D$ to $\ZT$.}
\end{center}
\begin{itemize}
    \item \textbf{Step 1.}~Construct the graph $G = (V,E)$ and
    the matrix $\mat{B} \in \bbQ^{E \times E}$
    from $\mat{K}^1, \ldots, \mat{K}^n$
    according to the procedure described in the beginning of the proof (\cref{eq:def-G-V,eq:def-G-E,eq:def-G-El,eq:def-G-Er,eq:def-B}).
    \item \textbf{Step 2.}~Determine if there exists a subset $S \subseteq E $ such that
    $ S \in \calT $, $\Er \subseteq S$, and $\det(\mat{B}_{S}) > 0$ by matroid intersection in polynomial time \cite{edmonds1970submodular}.
    If no such a subset has been found,
    then declare that ``$D(\mat{K}^1, \ldots, \mat{K}^n) = 0$,'' and otherwise, denote the subset found by $\tilde{S}$.
    \item \textbf{Step 3.}~Calculate the value of $x$ according to \cref{eq:x-st-fpras}, which requires polynomial time in the input size and $\epsilon^{-1}$ because
    the size of $\mat{B}$ is bounded by a polynomial in
    the size of $\mat{K}^1, \ldots, \mat{K}^n$ and
    the determinant can be computed in polynomial time
    by Gaussian elimination
    \cite{edmonds1967systems,schrijver1998theory}.
    \item \textbf{Step 4.}~Call an oracle for $\ZT$ on
    $\mat{B} \circ \mat{X}$ (which is positive semi-definite) and $G$
    with error tolerance $\delta = \epsilon/2$ to obtain
    an $\rme^{\epsilon/2}$-approximation to
    $\ZZ(x) = \ZT(\mat{B} \circ \mat{X}, G)$,
    which will be denoted by $\hat{\ZZ}$.
    \item \textbf{Step 5.}~Output $\displaystyle\frac{\hat{\ZZ}}{x^{2m}m!}$ as an estimate  for
    $D(\mat{K}^1, \ldots, \mat{K}^n)$.
\end{itemize}
\end{oframed}
By \cref{eq:st-fpras-bound},
if the oracle meets the specification for an FPRAS for $\ZT$,
then the output of the AP-reduction described above satisfies that
\begin{align*}
    \rme^{-\epsilon} \cdot D(\mat{K}^1, \ldots, \mat{K}^n) \leq \frac{\hat{\ZZ}}{x^{2m} m!} \leq \rme^{\epsilon} \cdot D(\mat{K}^1, \ldots, \mat{K}^n)
\end{align*}
with probability at least $\frac{3}{4}$.
Therefore, the AP-reduction meets the specification for an FPRAS for $D$,
which completes the proof.
\qed

\section{Proof of \cref{thm:fo-fpras}}
\label{sec:fo-fpras}
We construct an AP-reduction from the mixed discriminant $D$ to $\ZF$.
Suppose we have an FPRAS for $\ZF$; i.e.,
we can approximate $\ZF$ within a factor of $\rme^{\delta}$ in polynomial time.
Let $\epsilon \in (0,1)$ be an error tolerance for $D$,
i.e., we are asked to estimate $D$ within a factor of $\rme^{\epsilon}$.

Let $\mat{K}^1, \ldots, \mat{K}^n$ be $n$ positive semi-definite matrices in $\bbQ^{n \times n}$, and let $m = n^2$.
As with the proof of \cref{thm:st-fpras}, according to \cite[Proof of Lemma 12]{celis2017complexity},
we construct an $m \times m$ positive semi-definite matrix $\mat{A}$ and
an equal-sized partition of $[m]$, denoted
$P_1, P_2, \ldots, P_n$,
such that 
$\sum_{S \in \calC} \det(\mat{A}_{S}) = m! \; D(\mat{K}^1, \ldots, \mat{K}^n)$,
where
$ \calC = \{S \in {[m] \choose n} \mid |S \cap P_i| = 1 \; \forall i \in [n]\} $.
We then construct a simple, undirected graph $G=(V,E)$ according to
\cref{eq:def-G-V,eq:def-G-E,eq:def-G-El,eq:def-G-Er} and
a matrix $\mat{B} \in \bbQ^{E \times E}$
according to \cref{eq:def-B}.
Recall that $|V| = n+m+1$, $|E|=2m$, and $\mat{B}$ is positive semi-definite.
Let $\calF$ be the family of the edge sets of all forests of $G$.
We rephrase \cref{claim:thm:st-fpras} in the proof of \cref{thm:st-fpras} as follows.

\begin{claim*}
Let $S$ be an edge set with $\Er \subseteq S \subseteq E$.
Then, $S$ induces a forest of $G$ and $|S \cap \El| = n$ if and only if 
the set $\{ j \in [m] \mid (v_i, w_{i,j}) \in S \cap \El \}$ is contained in $\calC$.
\end{claim*}
The following equality is a direct consequence of the claim.
\begin{align}
\label{eq:sum-fo}
\sum_{\substack{S \in \calF: \Er \subseteq S \\ |S \cap \El| = n}} \det(\mat{B}_S)
= \sum_{S \in \calC} \det(\mat{A}_S).
\end{align}

Introduce two positive rational numbers $x, y \in \bbQ$ (whose values will be determined later) and define a matrix $\mat{X} \in \bbQ^{E \times E}$ depending
on $x$ and $y$ as follows:
\begin{align*}
    X_{i,j} =
    \begin{cases}
    y^2 & \text{if } i, j \in \El, \\
    xy & \text{if } i \in \El, j \in \Er, \\
    xy & \text{if } j \in \El, i \in \Er, \\
    x^2 & \text{if } i,j \in \Er.
    \end{cases}
\end{align*}
Since $\mat{X}$ is positive semi-definite for
any $x,y>0$ (its eigenvalues are $0$ and $m(x^2+y^2)$),
so is $\mat{B} \circ \mat{X}$ by the Schur product theorem \cite{schur1911bemerkungen}.
It is easy to observe that for each $S \subseteq E$,
\begin{align*}
\det((\mat{B} \circ \mat{X})_S) = x^{2|S \cap \Er|} y^{2|S \cap \El|} \det(\mat{B}_S).    
\end{align*}
We define a \emph{bivariate} polynomial in $x$ and $y$ as follows:
\begin{align*}
\ZZ(x,y) = \ZF(\mat{B} \circ \mat{X}, G) = \sum_{S \in \calF} x^{2|S \cap \Er|} y^{2|S \cap \El|} \det(\mat{B}_S).    
\end{align*}
By \cref{eq:sum-fo}, the coefficient of $x^{2m}y^{2n}$ in $\ZZ(x,y)$ is exactly the desired value,
i.e., $m! \; D(\mat{K}^1, \ldots, \mat{K}^n)$.

We now identify the values of $x$ and $y$ such that 
$\ZZ(x,y)$ is sufficiently close to $m! \; D(\mat{K}^1, \ldots, \mat{K}^n)$.
We first verify whether there exists
a subset $S \in \calF$ with $\Er \subseteq S$ and $|\El \cap S| = n$ such that $\det(\mat{B}_S) > 0$
because otherwise,
we can safely declare that \cref{eq:sum-fo} is $0$;
i.e., $D(\mat{K}^1, \ldots, \mat{K}^n)$ is $0$.
Such a subset can be found (if exists) by matroid intersection in a similar manner to that in the proof of \cref{thm:st-fpras} and denote
the subset found by $\tilde{S}$.
We fix values of $x$ and $y$ as follows:
\begin{align}
\label{eq:x-fo-fpras}
y = \frac{\det(\mat{B} + \mat{I})}{\det(\mat{B}_{\tilde{S}})} \frac{4}{\epsilon} \;\;\text{ and }\;\;
x = \frac{\det(\mat{B}+\mat{I})}{\det(\mat{B}_{\tilde{S}})} y^{2m-2n} \frac{4}{\epsilon}.
\end{align}
Notice that $x>1$ and $y>1$.
Hence, each monomial $x^{2|S \cap \Er|} y^{2|S \cap \El|} \det(\mat{B}_S)$
associated with $S \in \calF$
fits into one of the following three cases:
\begin{itemize}
    \item \textbf{Case 1.}~If $\Er \subseteq S, |S \cap \El| = n$:
    $x^{2|S \cap \Er|} y^{2|S \cap \El|} \leq x^{2m}y^{2n}$;
    \item \textbf{Case 2.}~If $\Er \subseteq S, |S \cap \El| < n$:
    $x^{2|S \cap \Er|} y^{2|S \cap \El|} \leq x^{2m}y^{2n-2}$;
    \item \textbf{Case 3.}~If $\Er \not \subseteq S$:
    $x^{2|S \cap \Er|} y^{2|S \cap \El|} \leq x^{2m-2}y^{2m}$.
\end{itemize}
Define $\sigma$ as follows:
\begin{align*}
    \sigma = m! \; D(\mat{K}^1, \ldots, \mat{K}^n) = \sum_{\substack{S \in \calF: \Er \subseteq S \\ |S \cap \El| = n}} \det(\mat{B}_S).
\end{align*}
Observing that $\sigma \geq \det(\mat{B}_{\tilde{S}})$,
we now bound $\ZZ(x,y) $ from above as follows:
\begin{align*}
& \sum_{S \in \calF} x^{2|S \cap \Er|} y^{2|S \cap \El|} \det(\mat{B}_S) \\
& \leq \sum_{\substack{S \in \calF: \Er \subseteq S \\ |S \cap \El| = n}} x^{2m} y^{2n} \det(\mat{B}_S)
+ \sum_{\substack{S \in \calF: \Er \subseteq S \\ |S \cap \El| \leq n-1}} x^{2m} y^{2n-2} \det(\mat{B}_S)  + \sum_{S \in \calF: \Er \not \subseteq S} x^{2m-2} y^{2m} \det(\mat{B}_S) \\
& = x^{2m} y^{2n} \sigma
\left[1 + \frac{\sum\limits_{\substack{S \in \calF: \Er \subseteq S \\ |S \cap \El| \leq n-1}} \det(\mat{B}_S)}{\sum\limits_{\substack{S \in \calF: \Er \subseteq S \\ |S \cap \El| = n}} \det(\mat{B}_S)}\frac{1}{y^2} + \frac{\sum\limits_{S \in \calF: \Er \not \subseteq S} \det(\mat{B}_S)}{\sum\limits_{\substack{S \in \calF: \Er \subseteq S \\ |S \cap \El| = n}} \det(\mat{B}_S)}\frac{y^{2m-2n}}{x^2}
\right] \\
& \leq x^{2m}y^{2n} \sigma
\left[
1 + \frac{\det(\mat{B}_{\tilde{S}})}{\sigma} \frac{\sum\limits_{\substack{S \in \calF: \Er \subseteq S \\ |S \cap \El| \leq n-1}} \det(\mat{B}_S)}{\det(\mat{B}+\mat{I})}\frac{\epsilon}{4}\frac{1}{y} + \frac{\det(\mat{B}_{\tilde{S}})}{\sigma}\frac{\sum\limits_{S \in \calF: \Er \not \subseteq S} \det(\mat{B}_S)}{\det(\mat{B}+\mat{I})} \frac{\epsilon}{4}\frac{1}{x}
\right] \\
& \leq x^{2m}y^{2n} \left(1+\frac{\epsilon}{2}\right) \sigma \\
& \leq x^{2m}y^{2n}\rme^{\epsilon/2} m! \; D(\mat{K}^1, \ldots, \mat{K}^n).
\end{align*}
Since $\ZZ(x,y) \geq x^{2m} y^{2n}m! \; D(\mat{K}^1, \ldots, \mat{K}^n)$,
we have that
\begin{align}
\label{eq:fo-fpras-bound}
    D(\mat{K}^1, \ldots, \mat{K}^n) \leq \frac{\ZZ(x,y)}{x^{2m} y^{2n} m!} \leq \rme^{\epsilon/2} \cdot D(\mat{K}^1, \ldots, \mat{K}^n).
\end{align}

We are finally ready to describe the AP-reduction from
the mixed discriminant $D$ to $\ZF$.
\begin{oframed}
\begin{center}
    \textbf{AP-reduction from $D$ to $\ZF$.}
\end{center}
\begin{itemize}
    \item \textbf{Step 1.}~Construct the graph $G = (V,E)$ and
    the matrix $\mat{B} \in \bbQ^{E \times E}$
    from $\mat{K}^1, \ldots, \mat{K}^n$
    according to the procedure described in the beginning of the proof (\cref{eq:def-G-V,eq:def-G-E,eq:def-G-El,eq:def-G-Er,eq:def-B}).
    
    \item \textbf{Step 2.}~Determine if there exists a subset $S \subseteq E$ such that
    $S \in \calF$, $\Er \subseteq S$, $|\El \cap S| = n$, and $\det(\mat{B}_{S}) > 0$ by matroid intersection in polynomial time \cite{edmonds1970submodular}.
    If no such a subset has been found, then declare that 
    ``$D(\mat{K}^1, \ldots, \mat{K}^n)$ = 0,'' and otherwise
    denote the subset found by $\tilde{S}$.
    \item \textbf{Step 3.}~Calculate the values of $x$ and $y$ according to \cref{eq:x-fo-fpras}, which requires polynomial time in the input size and $\epsilon^{-1}$ because
    the size of $\mat{B}$ is bounded by a polynomial in the size of $\mat{K}^1, \ldots, \mat{K}^n$ and
    the determinant can be computed in polynomial time
    by Gaussian elimination     \cite{edmonds1967systems,schrijver1998theory}.
    \item \textbf{Step 4.}~Call an oracle for $\ZF$ on
    $\mat{B} \circ \mat{X}$ and $G$ with error tolerance $\delta = \epsilon/2$ to obtain
    an $\rme^{\epsilon/2}$-approximation to
    $\ZZ(x,y) = \ZF(\mat{B} \circ \mat{X}, G)$,
    which will be denoted by $\hat{\ZZ}$.
    \item \textbf{Step 5.}~Output $\displaystyle\frac{\hat{\ZZ}}{x^{2m}y^{2n}m!}$ as an estimate for
    $D(\mat{K}^1, \ldots, \mat{K}^n)$.
\end{itemize}
\end{oframed}

By \cref{eq:fo-fpras-bound},
if the oracle meets the specification for an FPRAS for $\ZF$,
then the output of the AP-reduction described above satisfies that
\begin{align*}
    \rme^{-\epsilon} \cdot D(\mat{K}^1, \ldots, \mat{K}^n) \leq \frac{\hat{\ZZ}}{x^{2m}y^{2n}m!} \leq \rme^{\epsilon} \cdot D(\mat{K}^1, \ldots, \mat{K}^n)
\end{align*}
with probability at least $\frac{3}{4}$.
Therefore, the AP-reduction meets the specification of an FPRAS for $D$,
which completes the proof.
\qed

\bibliographystyle{abbrvnat}
\bibliography{main}

\begin{thebibliography}{29}
\providecommand{\natexlab}[1]{#1}
\providecommand{\url}[1]{\texttt{#1}}
\expandafter\ifx\csname urlstyle\endcsname\relax
  \providecommand{\doi}[1]{doi: #1}\else
  \providecommand{\doi}{doi: \begingroup \urlstyle{rm}\Url}\fi

\bibitem[Affandi et~al.(2014)Affandi, Fox, Adams, and
  Taskar]{affandi2014learning}
R.~H. Affandi, E.~B. Fox, R.~P. Adams, and B.~Taskar.
\newblock Learning the parameters of determinantal point process kernels.
\newblock In \emph{ICML}, pages 1224--1232, 2014.

\bibitem[Andrzejak(1998)]{andrzejak1998algorithm}
A.~Andrzejak.
\newblock An algorithm for the {Tutte} polynomials of graphs of bounded
  treewidth.
\newblock \emph{Discrete Math.}, 190\penalty0 (1-3):\penalty0 39--54, 1998.

\bibitem[Annan(1994)]{annan1994randomised}
J.~D. Annan.
\newblock A randomised approximation algorithm for counting the number of
  forests in dense graphs.
\newblock \emph{Comb. Probab. Comput.}, 3:\penalty0 273--283, 1994.

\bibitem[Bodlaender(1998)]{bodlaender1998partial}
H.~L. Bodlaender.
\newblock A partial {\it k}-arboretum of graphs with bounded treewidth.
\newblock \emph{Theor. Comput. Sci.}, 209\penalty0 (1--2):\penalty0 1--45,
  1998.

\bibitem[Borodin and Rains(2005)]{borodin2005eynard}
A.~Borodin and E.~M. Rains.
\newblock {Eynard-Mehta} theorem, {Schur} process, and their {Pfaffian}
  analogs.
\newblock \emph{J. Stat. Phys.}, 121\penalty0 (3--4):\penalty0 291--317, 2005.

\bibitem[Celis et~al.(2017)Celis, Deshpande, Kathuria, Straszak, and
  Vishnoi]{celis2017complexity}
L.~E. Celis, A.~Deshpande, T.~Kathuria, D.~Straszak, and N.~K. Vishnoi.
\newblock On the complexity of constrained determinantal point processes.
\newblock In \emph{APPROX/RANDOM}, pages 36:1--36:22, 2017.

\bibitem[Celis et~al.(2018)Celis, Keswani, Straszak, Deshpande, Kathuria, and
  Vishnoi]{celis2018fair}
L.~E. Celis, V.~Keswani, D.~Straszak, A.~Deshpande, T.~Kathuria, and N.~K.
  Vishnoi.
\newblock Fair and diverse {DPP}-based data summarization.
\newblock In \emph{ICML}, pages 715--724, 2018.

\bibitem[Courcelle et~al.(2001)Courcelle, Makowsky, and
  Rotics]{courcelle2001fixed}
B.~Courcelle, J.~A. Makowsky, and U.~Rotics.
\newblock On the fixed parameter complexity of graph enumeration problems
  definable in monadic second-order logic.
\newblock \emph{Discrete Appl. Math.}, 108\penalty0 (1-2):\penalty0 23--52,
  2001.

\bibitem[Cygan et~al.(2015)Cygan, Fomin, Kowalik, Lokshtanov, Marx, Pilipczuk,
  Pilipczuk, and Saurabh]{cygan2015parameterized}
M.~Cygan, F.~V. Fomin, {\L}.~Kowalik, D.~Lokshtanov, D.~Marx, M.~Pilipczuk,
  M.~Pilipczuk, and S.~Saurabh.
\newblock \emph{Parameterized Algorithms}.
\newblock Springer, 2015.

\bibitem[Dagum and Luby(1992)]{dagum1992approximating}
P.~Dagum and M.~Luby.
\newblock Approximating the permanent of graphs with large factors.
\newblock \emph{Theor. Comput. Sci.}, 102\penalty0 (2):\penalty0 283--305,
  1992.

\bibitem[Dyer et~al.(2004)Dyer, Goldberg, Greenhill, and
  Jerrum]{dyer2004relative}
M.~Dyer, L.~A. Goldberg, C.~Greenhill, and M.~Jerrum.
\newblock The relative complexity of approximate counting problems.
\newblock \emph{Algorithmica}, 38\penalty0 (3):\penalty0 471--500, 2004.

\bibitem[Edmonds(1967)]{edmonds1967systems}
J.~Edmonds.
\newblock Systems of distinct representatives and linear algebra.
\newblock \emph{J. Res. Natl. Bur. Stand.}, 71B:\penalty0 241--245, 1967.

\bibitem[Edmonds(1970)]{edmonds1970submodular}
J.~Edmonds.
\newblock Submodular functions, matroids, and certain polyhedra.
\newblock \emph{Combinatorial Structures and Their Applications}, pages 69--87,
  1970.

\bibitem[Fung et~al.(2011)Fung, Hariharan, Harvey, and
  Panigrahi]{fung2011general}
W.~S. Fung, R.~Hariharan, N.~J.~A. Harvey, and D.~Panigrahi.
\newblock A general framework for graph sparsification.
\newblock In \emph{STOC}, pages 71--80, 2011.

\bibitem[Gong et~al.(2014)Gong, Chao, Grauman, and Sha]{gong2014diverse}
B.~Gong, W.~Chao, K.~Grauman, and F.~Sha.
\newblock Diverse sequential subset selection for supervised video
  summarization.
\newblock In \emph{NIPS}, pages 2069--2077, 2014.

\bibitem[Gurvits(2005)]{gurvits2005complexity}
L.~Gurvits.
\newblock On the complexity of mixed discriminants and related problems.
\newblock In \emph{MFCS}, pages 447--458, 2005.

\bibitem[Hayashi et~al.(2016)Hayashi, Akiba, and Yoshida]{hayashi2016efficient}
T.~Hayashi, T.~Akiba, and Y.~Yoshida.
\newblock Efficient algorithms for spanning tree centrality.
\newblock In \emph{IJCAI}, pages 3733--3739, 2016.

\bibitem[Jerrum et~al.(2004)Jerrum, Sinclair, and Vigoda]{jerrum2004polynomial}
M.~Jerrum, A.~Sinclair, and E.~Vigoda.
\newblock A polynomial-time approximation algorithm for the permanent of a
  matrix with nonnegative entries.
\newblock \emph{J. ACM}, 51\penalty0 (4):\penalty0 671--697, 2004.

\bibitem[Jerrum et~al.(1986)Jerrum, Valiant, and Vazirani]{jerrum1986random}
M.~R. Jerrum, L.~G. Valiant, and V.~V. Vazirani.
\newblock Random generation of combinatorial structures from a uniform
  distribution.
\newblock \emph{Theor. Comput. Sci.}, 43:\penalty0 169--188, 1986.

\bibitem[Kirchhoff(1847)]{kirchhoff1847ueber}
G.~Kirchhoff.
\newblock Ueber die aufl{\"o}sung der gleichungen, auf welche man bei der
  untersuchung der linearen vertheilung galvanischer str{\"o}me gef{\"u}hrt
  wird.
\newblock \emph{Annalen der Physik}, 148\penalty0 (12):\penalty0 497--508,
  1847.

\bibitem[Kulesza and Taskar(2011)]{kulesza2011kdpps}
A.~Kulesza and B.~Taskar.
\newblock $k$-{DPPs}: Fixed-size determinantal point processes.
\newblock In \emph{ICML}, pages 1193--1200, 2011.

\bibitem[Kulesza and Taskar(2012)]{kulesza2012determinantal}
A.~Kulesza and B.~Taskar.
\newblock Determinantal point processes for machine learning.
\newblock \emph{Found. Trends Mach. Learn.}, 5\penalty0 (2--3):\penalty0
  123--286, 2012.

\bibitem[Macchi(1975)]{macchi1975coincidence}
O.~Macchi.
\newblock The coincidence approach to stochastic point processes.
\newblock \emph{Adv. Appl. Probab.}, 7\penalty0 (1):\penalty0 83--122, 1975.

\bibitem[Noble(1998)]{noble1998evaluating}
S.~D. Noble.
\newblock Evaluating the {Tutte} polynomial for graphs of bounded tree-width.
\newblock \emph{Comb. Probab. Comput.}, 7\penalty0 (3):\penalty0 307--321,
  1998.

\bibitem[Robertson and Seymour(1986)]{robertson1986graph}
N.~Robertson and P.~D. Seymour.
\newblock Graph minors. {II.} {A}lgorithmic aspects of tree-width.
\newblock \emph{J. Algorithms}, 7\penalty0 (3):\penalty0 309--322, 1986.

\bibitem[Schrijver(1999)]{schrijver1998theory}
A.~Schrijver.
\newblock \emph{Theory of Linear and Integer Programming}.
\newblock Wiley--Interscience Series in Discrete Mathematics and Optimization.
  Wiley, 1999.

\bibitem[Schur(1911)]{schur1911bemerkungen}
I.~Schur.
\newblock Bemerkungen zur theorie der beschr{\"a}nkten bilinearformen mit
  unendlich vielen ver{\"a}nderlichen.
\newblock \emph{Journal f{\"u}r die reine und angewandte Mathematik},
  1911\penalty0 (140):\penalty0 1--28, 1911.

\bibitem[Valiant(1979)]{valiant1979complexity}
L.~G. Valiant.
\newblock The complexity of computing the permanent.
\newblock \emph{Theor. Comput. Sci.}, 8\penalty0 (2):\penalty0 189--201, 1979.

\bibitem[Vertigan and Welsh(1992)]{vertigan1992computational}
D.~L. Vertigan and D.~J.~A. Welsh.
\newblock The computational complexity of the {Tutte} plane: The bipartite
  case.
\newblock \emph{Comb. Probab. Comput.}, 1\penalty0 (2):\penalty0 181--187,
  1992.

\end{thebibliography}

\end{document}